\documentclass[11pt]{article}
\usepackage[final]{acl}
\usepackage{microtype}
\usepackage{float}
\usepackage{times}
\usepackage{latexsym}
\usepackage[T1]{fontenc}
\usepackage[utf8]{inputenc}
\usepackage{microtype}
\usepackage{graphicx}
\usepackage{amsmath}
\usepackage{amssymb}
\usepackage{booktabs}
\usepackage{multirow}
\usepackage{xcolor}
\usepackage{algorithm}
\usepackage{algpseudocode}
\usepackage{subcaption}
\usepackage{tikz}
\usepackage{pgfplots}
\pgfplotsset{compat=1.18}
\usetikzlibrary{arrows.meta,positioning,shapes.geometric,fit,calc,decorations.markings}
\usepackage{enumitem}
\usepackage{pifont}
\newcommand{\cmark}{\ding{51}}
\newcommand{\xmark}{\ding{55}}

\title{RouteNLP: Closed-Loop LLM Routing with Conformal Cascading \\and Distillation Co-Optimization}

\author{
	Dongxin Guo$^{1}$, Jikun Wu$^{2}$, Siu Ming Yiu$^{1}$ \\
	$^{1}$The University of Hong Kong \quad $^{2}$Stellaris AI Limited \\
	\texttt{bettyguo@connect.hku.hk}, \texttt{hk950014@connect.hku.hk}, \\
	\texttt{smyiu@cs.hku.hk}
}

\begin{document}
\maketitle

\begin{abstract}
	Serving diverse NLP workloads with large language models is costly:
	at one enterprise partner, inference costs exceeded \$200K/month
	despite over 70\% of queries being routine tasks well within the
	capability of smaller models.
	We present \textsc{RouteNLP}, a closed-loop framework that routes
	queries across a tiered model portfolio to minimize cost while
	satisfying per-task quality constraints.
	The framework integrates three components:
	a difficulty-aware router with shared task-conditioned representations
	trained on preference data and quality signals;
	confidence-calibrated cascading that uses conformal prediction for
	distribution-free threshold initialization;
	and a distillation-routing co-optimization loop that clusters
	escalation failures, applies targeted knowledge distillation to cheaper
	models, and automatically retrains the router, yielding over twice the
	cost improvement of untargeted distillation.
	In an 8-week pilot deployment processing ${\sim}$5K queries/day
	at an enterprise customer-service division, \textsc{RouteNLP} reduced
	inference costs by 58\% while maintaining 91\% response acceptance and
	reducing p99 latency from 1,847\,ms to 387\,ms.
	On a six-task benchmark spanning finance, customer service, and legal
	domains, the framework achieves 40--85\% cost reduction while retaining
	96--100\% quality on structured tasks and 96--98\% on generation tasks,
	with human evaluation confirming that 74.5\% of routed generation
	outputs match or exceed frontier-model quality.
\end{abstract}

\section{Introduction}
\label{sec:intro}

LLMs have proliferated quickly, and enterprise teams now face a sharp tradeoff: models spanning orders of magnitude in cost and capability are available, from lightweight distilled models costing fractions of a cent to frontier models costing dollars per thousand tokens~\citep{chen2024frugalgpt}.
Financial institutions~\citep{wu2023bloomberggpt}, travel platforms~\citep{zagyva2025speed}, and customer service operations all face the same tension: \emph{how to deliver consistent quality while minimizing inference costs under strict latency constraints}.

This paper arose from a concrete production need. Working with an enterprise partner in financial services, we observed NLP serving costs exceeding \$200K/month, yet over 70\% of queries were routine tasks that did not require frontier model capabilities. Across enterprise domains (finance, customer service, legal), only 25--35\% of queries require frontier models, reflecting a heavy-tailed difficulty distribution that \textsc{RouteNLP} exploits (see Appendix~\ref{app:scenarios} for detailed deployment scenarios). For example, extracting standard entities from templated SEC filings is straightforward for small models, while summarizing novel regulatory guidance demands frontier capabilities; similarly, in customer service, over 70\% of queries are routine (order status, FAQ matching) where small models suffice.

Existing routing approaches have critical limitations for enterprise deployment: they are typically evaluated on single benchmarks~\citep{ong2025routellm,ding2024hybrid}, ignore production constraints such as latency SLAs~\citep{chen2024frugalgpt}, and most importantly decouple the router from the model portfolio, treating the set of available models as a fixed input rather than a learnable artifact. The recent unified framework of \citet{dekoninck2025unified} provides theoretical foundations, but a gap remains between academic frameworks and production systems.
\textsc{RouteNLP}'s central contribution is to break the fixed-portfolio assumption by closing the loop between routing failures and the portfolio itself: escalation logs are clustered, used to generate targeted distillation data, folded back into cheaper-tier models, after which the router and conformal thresholds are recalibrated. This change yields over twice the cost reduction of untargeted distillation at equal data volume (21.7\% vs.\ 9.4\%, \S\ref{sec:results}) and is, to our knowledge, the first such loop validated in a multi-week production deployment.

Our specific contributions are:
\begin{itemize}[nosep,leftmargin=*]
    \item A \textbf{closed-loop distillation-routing co-optimization loop} (\S\ref{sec:distill}) that clusters escalation failures, applies targeted knowledge distillation to cheaper tiers, and automatically retrains the router. At equal data volume, this yields over $2\times$ the cost gain of untargeted distillation.
    \item A \textbf{multi-task difficulty-aware router} (\S\ref{sec:router}) with shared task-conditioned representations, trained jointly on preference data and per-task quality signals so that a single encoder can learn task-dependent difficulty patterns.
    \item \textbf{Confidence-calibrated cascading} (\S\ref{sec:cascade}) that uses conformal risk control to initialize thresholds in a distribution-free manner. We are explicit about its caveats: the guarantee is marginal rather than per-query, and distribution shift can violate it.
    \item A \textbf{six-task multi-domain benchmark} (finance, customer service, legal) and an \textbf{8-week pilot deployment} ($\sim$5K queries/day) that validates the simulation predictions to within a 4-point gap on cost reduction (62\% benchmark, 58\% pilot; \S\ref{sec:experiments}--\S\ref{sec:deployment}).
\end{itemize}

\section{Related Work}
\label{sec:related}

\paragraph{LLM Routing and Cascading.}
Model routing was pioneered by \citet{chen2020frugalml} and extended to LLMs by \citet{chen2024frugalgpt}. Subsequent work trained routers on human preference data~\citep{ong2025routellm}, introduced tunable quality thresholds~\citep{ding2024hybrid}, formulated routing as a POMDP~\citep{aggarwal2024automix}, and used meta-modeling for per-query performance prediction~\citep{sakota2024forc,shnitzer2023routing,pan2024metallm}. RouterBench~\citep{hu2024routerbench} established systematic evaluation. \citet{dekoninck2025unified} showed that optimal serving lies on a continuum between pure routing and cascading; \textsc{RouteNLP}'s design is consistent with this analysis. Industry systems such as Microsoft's Copilot routing~\citep{wang2024copilot} demonstrate production adoption. For cascading, \citet{varshney2022cascading} demonstrated up to 88.9\% savings, \citet{gupta2024token} proposed token-level uncertainty for deferral, and \citet{yue2024mixture} combined cascading with diverse reasoning. Our work extends this literature to multi-task enterprise settings with SLA awareness and distillation co-optimization. A detailed feature comparison with prior systems is in Appendix~\ref{app:comparison}.

\paragraph{Uncertainty, Distillation, and Efficient Serving.}
Principled deferral requires reliable confidence estimates; conformal prediction~\citep{angelopoulos2024conformal,quach2024conformal} and semantic entropy~\citep{kuhn2023semantic} provide calibrated or distribution-free approaches. Knowledge distillation~\citep{hinton2015distilling,sanh2019distilbert} creates capable small models as routing targets; our co-optimization loop is conceptually related to Self-Refine~\citep{madaan2024selfrefine} and active learning, but operates at the system level across multiple models. Production routing depends on efficient infrastructure: vLLM~\citep{kwon2023vllm}, continuous batching~\citep{yu2022orca}, speculative decoding~\citep{leviathan2023speculative,cai2024medusa}, S-LoRA~\citep{sheng2024slora}, and quantization~\citep{frantar2023gptq,lin2024awq}. MoE architectures~\citep{shazeer2017moe,jiang2024mixtral} provide an intra-model analogue to inter-model routing.

\paragraph{Position of \textsc{RouteNLP}.}
The literature above establishes routing, cascading, conformal calibration, and distillation as well-studied building blocks. The gap that motivates our work is that no prior system combines all four into a single pipeline that treats the model portfolio itself as a learned artifact rather than a fixed input. Table~\ref{tab:comparison} (Appendix~\ref{app:comparison}) makes this gap concrete along four axes (\emph{multi-task} evaluation, \emph{formal calibration}, \emph{co-optimization}, and \emph{SLA awareness}) that jointly characterize what an industrial routing system must offer. The next section presents the framework that fills this gap.

\section{The RouteNLP Framework}
\label{sec:method}

\begin{figure*}[t]
	\centering
	\begin{tikzpicture}[
		box/.style={rectangle, draw, rounded corners=3pt, minimum height=0.9cm, minimum width=2.0cm, font=\small, align=center},
		tier/.style={rectangle, draw, rounded corners=3pt, minimum height=0.8cm, minimum width=2.4cm, font=\footnotesize, align=center},
		check/.style={circle, draw, inner sep=0pt, font=\scriptsize, minimum size=1.15cm},
		arrow/.style={-{Stealth[length=2.5mm]}, thick},
		dasharrow/.style={-{Stealth[length=2.5mm]}, thick, dashed},
		lbl/.style={font=\scriptsize, text=gray},
		]
		
		\def\colInput{0}
		\def\colRouter{3.2}
		\def\colTier{7.2}
		\def\colCheck{10.8}
		\def\colOutput{14.0}
		
		\def\rowA{2.4}
		\def\rowB{0.8}
		\def\rowC{-0.8}
		\def\rowD{-2.4}
		
		\node[box, fill=blue!10] (input) at (\colInput, 0) {Query $x$\\Task $t$};
		\node[box, fill=orange!15] (router) at (\colRouter, 0) {Difficulty-Aware\\Router $r_\theta$};
		\node[tier, fill=green!20] (t1) at (\colTier, \rowA) {T1: DistilBERT\\\$0.01/1K};
		\node[tier, fill=green!10] (t2) at (\colTier, \rowB) {T2: Mistral-7B\\\$0.10/1K};
		\node[tier, fill=yellow!15] (t3) at (\colTier, \rowC) {T3: Mixtral\\\$0.80/1K};
		\node[tier, fill=red!15]    (t4) at (\colTier, \rowD) {T4: GPT-4-Turbo\\\$8.00/1K};
		\node[check] (u1) at (\colCheck, \rowA) {$u \!\leq\! \delta_{1,t}$?};
		\node[check] (u2) at (\colCheck, \rowB) {$u \!\leq\! \delta_{2,t}$?};
		\node[check] (u3) at (\colCheck, \rowC) {$u \!\leq\! \delta_{3,t}$?};
		\node[box, fill=blue!10] (output) at (\colOutput, 0) {Response\\$y$};
		
		\draw[arrow] (input) -- (router);
		\draw[arrow] (router) -- (t1.west);
		\draw[arrow] (router) -- (t2.west);
		\draw[arrow] (router) -- (t3.west);
		\draw[arrow] (router) -- (t4.west);
		\draw[arrow] (t1.east) -- (u1.west);
		\draw[arrow] (t2.east) -- (u2.west);
		\draw[arrow] (t3.east) -- (u3.west);
		
		\draw[arrow, color=green!60!black] (u1.east) -- (output.north west);
		\draw[arrow, color=green!60!black] (u2.east) -- (output.west);
		\draw[arrow, color=green!60!black] (u3.east) -- (output.south west);
		
		\draw[arrow] (t4.east) -- (\colOutput, \rowD) -- (output.south);
		
		\draw[dasharrow, color=red!60] (u1.south west) -- (t2.east)
		node[pos=0.4, left, lbl] {esc.};
		\draw[dasharrow, color=red!60] (u2.south west) -- (t3.east)
		node[pos=0.4, left, lbl] {esc.};
		\draw[dasharrow, color=red!60] (u3.south west) -- (t4.east)
		node[pos=0.4, left, lbl] {esc.};
		
		\node[box, fill=purple!10] (coopt) at (\colRouter, -3.6) {Co-Optimization\\Loop (\S\ref{sec:distill})};
		\draw[dasharrow, color=purple!60]
		(output.south) -- (\colOutput, -3.57) -- (coopt.east)
		node[pos=0.5, above, lbl] {failure logs};
		\draw[dasharrow, color=purple!60]
		(coopt.north) -- (router.south)
		node[midway, right, lbl] {updated models};
		
	\end{tikzpicture}
	\caption{\textsc{RouteNLP} architecture. The router assigns queries to model tiers; the cascade escalates uncertain responses (dashed red) with conformally calibrated thresholds; the co-optimization loop (dashed purple) improves cheaper models via targeted distillation on failure clusters.}
	\label{fig:architecture}
\end{figure*}
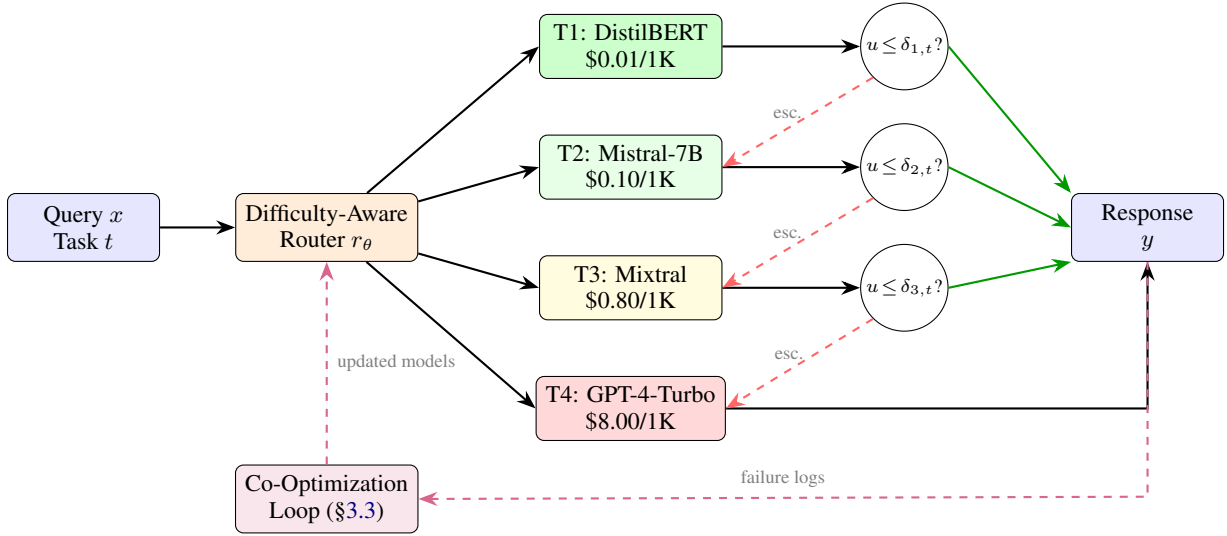

The system operates over a model portfolio $\mathcal{M} = \{m_1, \ldots, m_K\}$ ordered by cost $c_1 < \cdots < c_K$.
Given a query $x$ with task type $t$, let $k^*(x)$ denote the final tier handling $x$ (accounting for cascading). The system minimizes expected cost subject to a quality constraint:
\begin{equation}
\min_{r_\theta} \; \mathbb{E}_{x}\!\left[\textstyle\sum_{k=1}^{k^*(x)} c_{k,t}\right] \;\; \text{s.t.} \;\; \mathbb{E}_{x}[q(m_{k^*(x)}, x)] \geq \tau_t
\label{eq:objective}
\end{equation}
where $r_\theta$ is the learned routing policy, $q(m, x)$ is quality, and $\tau_t$ is the task-specific threshold. The cost sums over \emph{all} tiers attempted for cascaded queries. In practice, we approximate this via the composite loss in Eq.~\ref{eq:loss}.
Figure~\ref{fig:architecture} provides an overview.

\subsection{Difficulty-Aware Router}
\label{sec:router}

The router $r_\theta(x, t) \in \{1, \ldots, K\}$ predicts the cheapest model producing acceptable quality for query $x$ of task type $t$.
We train a lightweight DistilBERT-based classifier with a multi-task head that jointly predicts difficulty level and minimum model tier.
DistilBERT serves here as a \emph{router} (a difficulty classifier over the input query), not as a generator: T2--T4 are decoder-style language models that handle generation. For the structured tasks (NER, intent classification, clause extraction), the same DistilBERT additionally serves as the T1 generation model after task-specific fine-tuning, reusing the encoder's representations end-to-end; for generation tasks (Financial Summarization, CS Response), the router learns that cheaper tiers are insufficient and predominantly routes to T2--T4 (Table~\ref{tab:routing_dist}, Appendix~\ref{app:extended_results}).

Training labels are obtained by evaluating all queries on all models, labeling each $(x, m_k)$ pair with a binary quality indicator based on task-specific metrics (F1, ROUGE-L, or BERTScore $\geq \tau_t$; thresholds set per enterprise SLAs, with details in Appendix~\ref{app:router_details}).
The router learns to predict the cheapest model exceeding $\tau_t$.
For generation tasks, BERTScore serves as a training proxy; while it overstates absolute quality differences, the \emph{ranking} of model capabilities is preserved, making the router's binary sufficient/insufficient labels reliable (\S\ref{sec:human_eval} confirms 84--87\% agreement with human judgment).
Following \citet{ong2025routellm}, we augment with preference data from pairwise model comparisons; separately, following \citet{shnitzer2023routing}, we leverage benchmark performance to improve generalization.

Unlike prior routers that train separately per task, we use a shared encoder with task-specific projection heads.
A learned task embedding $\mathbf{e}_t \in \mathbb{R}^{64}$ is concatenated with the \texttt{[CLS]} representation, enabling task-dependent difficulty patterns.
Training uses a composite loss that is the differentiable surrogate of the constrained objective in Eq.~\ref{eq:objective}:
\begin{equation}
\mathcal{L} = \mathcal{L}_{\text{route}} + \lambda_c \cdot \mathcal{L}_{\text{cost}} + \lambda_q \cdot \mathcal{L}_{\text{quality}}
\label{eq:loss}
\end{equation}
Concretely, $\mathcal{L}_{\text{route}}$ replaces the combinatorial choice of $r_\theta$ with a tier-classification cross-entropy against tier labels derived from the all-pairs evaluation; $\mathcal{L}_{\text{cost}}$ encodes the cost objective (normalized by the maximum tier cost); and $\mathcal{L}_{\text{quality}}$ encodes the quality constraint as a hinge penalty on tiers predicted to fall below $\tau_t$. We set $\lambda_c\!=\!0.3$, $\lambda_q\!=\!0.5$ (sensitivity analysis in Appendix~\ref{app:sensitivity}).

\subsection{Confidence-Calibrated Cascading}
\label{sec:cascade}

When the router assigns a query to tier $k$, the system generates a response and evaluates a calibrated confidence score.
If confidence is insufficient, the query cascades to tier $k\!+\!1$.
Following \citet{gupta2024token}, we compute token-level uncertainty:
\begin{equation}
u(m_k, x) = \frac{1}{L} \sum_{i=1}^{L} \left(1 - p_{m_k}(y_i \mid y_{<i}, x)\right)
\label{eq:uncertainty}
\end{equation}
High uncertainty $u(m_k, x) > \delta_{k,t}$ triggers escalation, where $\delta_{k,t}$ is task- and tier-specific. For self-hosted tiers (T1--T3), $p_{m_k}(y_i \mid y_{<i}, x)$ is read directly from the decoder's softmax. For the API-served frontier tier (T4 = GPT-4-Turbo), Eq.~\ref{eq:uncertainty} is computed from the top-token log-probabilities exposed by the OpenAI \texttt{logprobs} parameter, which is sufficient to estimate the expected token-level uncertainty in our setting; comparable functionality is available in other major commercial LLM APIs.

We apply conformal risk control~\citep{angelopoulos2024conformal,bates2021distribution} to set thresholds: on 500 calibration examples per task and tier, we compute binary nonconformity scores $s_i = \mathbb{1}[q(m_k, x_i) < \tau_t]$ and set $\delta_{k,t}$ as the $\lceil (1\!-\!\alpha)(n_0\!+\!1) \rceil$-th quantile of uncertainty scores among correctly-handled examples. Under exchangeability, this provides the marginal guarantee:
\begin{equation}
\Pr\!\left[q(m_k, X) \!<\! \tau_t \;\wedge\; u(m_k, X) \!\leq\! \delta_{k,t}\right] \leq \alpha
\end{equation}
with $\alpha\!=\!0.05$. Three practical caveats apply: (1)~this is a marginal guarantee over the joint distribution (not per-query); (2)~it requires exchangeability, violated under distribution shift (see robustness analysis in \S\ref{sec:results}); (3)~calibration set size affects tightness. At 500 samples, the 95\% Wilson CI on the 4.2\% violation rate is [2.5\%, 6.6\%], indicating the true rate may marginally exceed the 5\% target. We recommend conformal thresholds as initialization supplemented by production monitoring (calibration details in Appendix~\ref{app:conformal}).
A cascaded query incurs cumulative cost: $\text{Cost}(x) = \sum_{k=1}^{k^*(x)} c_{k,t}$.

\subsection{Distillation--Routing Co-Optimization}
\label{sec:distill}

Unlike prior routing work treating the model portfolio as fixed, we iteratively improve cheaper models based on routing failure analysis, then retrain the router and recalibrate thresholds.
Algorithm~\ref{alg:coopt} provides the full procedure.

\begin{algorithm}[t]
\caption{Distillation--Routing Co-Optimization}
\label{alg:coopt}
\small
\begin{algorithmic}[1]
\Require Portfolio $\mathcal{M}$, data $\mathcal{D}$, threshold $\varepsilon = 0.005$
\State Train initial router $r_\theta^{(0)}$; calibrate $\{\delta_{k,t}^{(0)}\}$
\For{iteration $i = 1, 2, \ldots$}
    \State Collect escalation logs $\mathcal{F} \gets \{(x, t, k) : x \text{ escalated}\}$
    \State Extract router representations; PCA to 128-d; $k$-means ($k\!=\!10$) per task
    \State Rank clusters by size $\times$ avg.\ quality gap; select top-5/task
    \State Generate distillation data from frontier model on cluster exemplars
    \State Fine-tune $m_1, \ldots, m_{K-1}$ via SeqKD~\citep{kim2016sequence}
    \State Retrain router $r_\theta^{(i)}$; recalibrate $\{\delta_{k,t}^{(i)}\}$
    \If{$|\text{CostRatio}^{(i)} - \text{CostRatio}^{(i-1)}| < \varepsilon$}
        \textbf{break}
    \EndIf
\EndFor
\end{algorithmic}
\end{algorithm}

After each iteration, escalated queries are clustered using the router's hidden representations (PCA to 128-d, $k$-means with $k\!=\!10$ selected by silhouette score). Clusters are ranked by size $\times$ average quality gap, prioritizing large systematic failures.
The distillation dataset combines the top 30\% hardest failures with 20\% random in-distribution samples to prevent catastrophic forgetting. The loop converges in 2--3 iterations (details in Appendix~\ref{app:coopt_details}).

\section{Experimental Setup}
\label{sec:experiments}

\paragraph{Benchmark.}
We construct a six-task benchmark spanning three enterprise domains (Table~\ref{tab:benchmark}): Financial NER and Summarization (SEC EDGAR filings with expert annotations, inter-annotator F1: 0.93), CS Intent Classification and Response Generation (BANKING77~\citep{casanueva2020banking77} augmented with 3,000 enterprise-specific intents; reference responses by 3 agents, Krippendorff's $\alpha\!=\!0.78$), and Legal Clause Extraction and Risk Assessment (CUAD~\citep{hendrycks2021cuad} re-annotated by legal professionals, Cohen's $\kappa\!=\!0.83$). The benchmark adapts public datasets with enterprise annotations, enabling reproducibility while capturing domain difficulty patterns; pilot deployment validates directional consistency (\S\ref{sec:deployment}). Full provenance in Appendix~\ref{app:benchmark_details}.

\begin{table}[t]
	\centering
	\small
	\resizebox{\columnwidth}{!}{%
		\begin{tabular}{@{}llrrl@{}}
			\toprule
			\textbf{Domain} & \textbf{Task} & \textbf{Train} & \textbf{Test} & \textbf{Source} \\
			\midrule
			\multirow{2}{*}{Finance} & NER & 8,200 & 1,800 & EDGAR\textsuperscript{a} \\
			& Summarization & 5,400 & 1,200 & EDGAR\textsuperscript{a} \\
			\midrule
			\multirow{2}{*}{Cust.\ Svc.} & Intent Classif. & 12,000 & 2,600 & BANK77+\textsuperscript{b} \\
			& Response Gen. & 6,800 & 1,500 & BANK77+\textsuperscript{b} \\
			\midrule
			\multirow{2}{*}{Legal} & Clause Extract. & 4,600 & 1,000 & CUAD\textsuperscript{c} \\
			& Risk Assessment & 3,200 & 700 & CUAD\textsuperscript{c} \\
			\midrule
			\multicolumn{2}{l}{\textbf{Total}} & \textbf{40,200} & \textbf{8,800} & \\
			\bottomrule
		\end{tabular}%
	}
	\caption{Benchmark statistics. \textsuperscript{a}SEC EDGAR with expert annotations; \textsuperscript{b}BANKING77~\citep{casanueva2020banking77} augmented with enterprise intents; \textsuperscript{c}CUAD~\citep{hendrycks2021cuad} with re-annotation.}
	\label{tab:benchmark}
\end{table}

\paragraph{Model Portfolio.}
Four tiers spanning ${\sim}$800$\times$ cost range. T1: DistilBERT~\citep{sanh2019distilbert} fine-tuned per task (\$0.01/1K). T2: Mistral-7B-Instruct with LoRA~\citep{sheng2024slora} (\$0.10/1K). T3: Mixtral-8$\times$7B~\citep{jiang2024mixtral} with AWQ quantization~\citep{lin2024awq} (\$0.80/1K). T4: GPT-4-Turbo via API (\$8.00/1K avg). Open-source models served via vLLM~\citep{kwon2023vllm} with speculative decoding~\citep{leviathan2023speculative,cai2024medusa}.

\paragraph{Baselines.}
We compare against Always-T4 (quality upper bound), Always-T2 (cost-efficient), Random, Rule-Based (structured$\to$T1, generation$\to$T3), FrugalGPT~\citep{chen2024frugalgpt}, Hybrid LLM~\citep{ding2024hybrid}, RouteLLM~\citep{ong2025routellm}, and AutoMix~\citep{aggarwal2024automix}. Originally 2-model baselines (Hybrid LLM, RouteLLM) were extended to 4-tier settings by replacing binary routing heads with 4-class heads trained on identical data; we additionally evaluated faithful 2-tier (Binary-T2/T4) variants and report the better-performing Extended-4-Tier configurations in the main results. The Binary variants incurred 2.1--3.4$\times$ higher costs (Appendix~\ref{app:baselines}). FrugalGPT's cascade naturally extends to 4 tiers; AutoMix's POMDP was reformulated with 4 actions. \textbf{All baselines received identical training data, model portfolio access, and calibration sets as \textsc{RouteNLP}.} Detailed adaptation protocol in Appendix~\ref{app:baselines}.

\paragraph{Metrics.}
Task-specific quality (F1, ROUGE-L, BERTScore, accuracy); Quality Ratio and Cost Ratio relative to Always-T4 (using cumulative cascade costs); p99 latency under simulated production load; SLA violation rate. All experiments over 5 seeds with paired bootstrap significance tests.

\section{Results and Analysis}
\label{sec:results}

\subsection{Overall Cost--Quality Tradeoff}

Table~\ref{tab:main_results} presents main results. \textsc{RouteNLP} achieves 40--85\% cost reduction across tasks (62\% on simulated production traffic; Table~\ref{tab:pilot}) while retaining 96--100\% quality on structured tasks and 96--98\% on generation tasks.

\begin{table}[t]
\centering
\small
\setlength{\tabcolsep}{2.8pt}
\begin{tabular}{@{}lcccc@{}}
\toprule
 & \textbf{Quality} & \textbf{Cost} & \textbf{p99} & \textbf{SLA} \\
\textbf{System} & \textbf{Ratio}$\uparrow$ & \textbf{Ratio}$\downarrow$ & \textbf{(ms)}$\downarrow$ & \textbf{Viol.}$\downarrow$ \\
\midrule
Always-T4       & 1.000\tiny{$\pm$.000} & 1.000\tiny{$\pm$.000} & 1,847 & 38.2\% \\
Always-T2       & .891\tiny{$\pm$.003} & .013\tiny{$\pm$.000} & 142 & 0.1\% \\
Random          & .924\tiny{$\pm$.008} & .252\tiny{$\pm$.012} & 623 & 12.4\% \\
Rule-Based      & .943\tiny{$\pm$.002} & .198\tiny{$\pm$.003} & 524 & 8.9\% \\
FrugalGPT      & .967\tiny{$\pm$.004} & .284\tiny{$\pm$.009} & 986 & 21.3\% \\
Hybrid LLM     & .972\tiny{$\pm$.005} & .312\tiny{$\pm$.011} & 874 & 18.7\% \\
RouteLLM        & .969\tiny{$\pm$.004} & .246\tiny{$\pm$.008} & 841 & 17.2\% \\
AutoMix         & .958\tiny{$\pm$.006} & .231\tiny{$\pm$.010} & 1,124 & 24.6\% \\
\midrule
\textsc{RouteNLP}  & \textbf{.971}\tiny{$\pm$.004} & \textbf{.159}\tiny{$\pm$.006} & \textbf{387} & \textbf{2.3\%} \\
~~w/o cascade   & .952\tiny{$\pm$.005} & .134\tiny{$\pm$.005} & 298 & 1.8\% \\
~~w/o co-opt.  & .961\tiny{$\pm$.005} & .203\tiny{$\pm$.008} & 412 & 3.1\% \\
~~w/o task cond.& .964\tiny{$\pm$.004} & .187\tiny{$\pm$.007} & 395 & 2.7\% \\
\bottomrule
\end{tabular}
\caption{Main results (mean $\pm$ std, 5 seeds) volume-weighted across six tasks. Quality and Cost Ratios are computed relative to Always-T4. The p99 column is from the M/M/c queueing simulation under matched production load (Appendix~\ref{app:latency}); the pilot empirical latency is reported separately in \S\ref{sec:deployment}. Bottom: ablations.}
\label{tab:main_results}
\end{table}

The cost reduction over RouteLLM (0.159 vs.\ 0.246) is statistically significant ($p < 0.001$, paired bootstrap). The quality difference vs.\ Hybrid LLM (0.971 vs.\ 0.972) is not significant ($p = 0.82$), confirming comparable quality at 49\% lower cost. SLA violations drop from 17.2\% (RouteLLM) to 2.3\%, a 7.5$\times$ improvement.

\paragraph{Ablations.}
Removing the cascade reduces quality by 1.9 points (the cascade serves as a critical safety net for ambiguous queries; its lower cost is illusory since the quality drop is unacceptable for constrained deployments). Removing co-optimization increases cost by 28\% because small models cannot handle as many queries without targeted improvement. Removing task conditioning increases cost by 18\%, confirming task-aware difficulty prediction is important for multi-task settings. To isolate failure clustering's contribution, we compared against random distillation (same data volume, no failure analysis): targeted distillation reduces cost ratio from 0.203 to 0.159 (21.7\% reduction), versus 0.184 for random distillation (9.4\%). Failure clustering thus provides over twice the cost improvement.

\paragraph{Per-Task Analysis.}
Table~\ref{tab:per_task} shows per-task breakdowns.
Structured tasks (NER, Intent, Clause Extraction) achieve 78--85\% cost reduction with $<$1\% absolute quality loss. Generation tasks achieve 40--47\% savings with larger drops: CS Response shows a 2.7-point BERTScore drop and Financial Summarization a 1.8-point ROUGE-L drop. We address whether these automated metric drops correspond to perceived quality differences in \S\ref{sec:human_eval}.

\begin{table}[t]
\centering
\small
\setlength{\tabcolsep}{3pt}
\begin{tabular}{@{}lccccc@{}}
\toprule
\textbf{Task} & \textbf{Metric} & \textbf{T4} & \textbf{Ours} & \textbf{Retain} & \textbf{Cost$\downarrow$}\\
\midrule
Fin.\ NER & F1 & 94.2 & 93.8\tiny{$\pm$.3} & 99.6\% & 82\% \\
Fin.\ Summ. & R-L & 48.7 & 46.9\tiny{$\pm$.4} & 96.3\% & 47\% \\
CS Intent & F1 & 96.1 & 95.8\tiny{$\pm$.2} & 99.7\% & 85\% \\
CS Resp. & BS & 72.4 & 69.7\tiny{$\pm$.6} & 96.3\% & 42\% \\
Legal Cl. & F1 & 91.6 & 90.9\tiny{$\pm$.3} & 99.2\% & 78\% \\
Legal Risk & Acc & 88.3 & 86.1\tiny{$\pm$.5} & 97.5\% & 40\% \\
\bottomrule
\end{tabular}
\caption{Per-task quality. Retain = quality retention vs.\ Always-T4. BS = BERTScore, R-L = ROUGE-L. Std over 5 seeds.}
\label{tab:per_task}
\end{table}

\paragraph{Robustness and Threshold Sensitivity.}
The system degrades gracefully under distribution shift: difficulty shift increases cost to 0.214 while maintaining 96.3\% quality; domain shift raises coverage violations to 8.1\% (exceeding the 5\% target), indicating recalibration is needed for significant domain changes. Cost savings are robust across $\pm$10\% threshold variation (69--89\% savings); even at +5\% stricter thresholds, \textsc{RouteNLP}'s cost (0.218) remains below RouteLLM (0.308) and Hybrid LLM (0.381). Full robustness results and routing distributions are in Appendix~\ref{app:extended_results}.

\subsection{Co-Optimization Convergence}

Table~\ref{tab:coopt} shows convergence in 3 iterations.
After iteration~1, cost ratio drops from 0.203 to 0.178 as targeted distillation improves T1/T2 on failure clusters.
After iteration~3, cost ratio reaches 0.159 with quality stable at 0.971.
Each iteration progressively shifts queries from expensive to cheap tiers: approximately 6\% in iteration~1, 5\% in iteration~2, and 2\% in iteration~3 as remaining failures become harder to address.

\begin{table}[t]
\centering
\small
\begin{tabular}{@{}lcccc@{}}
\toprule
\textbf{Iter.} & \textbf{Quality$\uparrow$} & \textbf{Cost$\downarrow$} & \textbf{T1+T2} & \textbf{T4} \\
\midrule
0 (init.) & .961\tiny{$\pm$.005} & .203\tiny{$\pm$.008} & 68\% & 11\% \\
1 & .964\tiny{$\pm$.004} & .178\tiny{$\pm$.007} & 74\% & 8\% \\
2 & .969\tiny{$\pm$.004} & .163\tiny{$\pm$.006} & 79\% & 6\% \\
3 (final) & .971\tiny{$\pm$.004} & .159\tiny{$\pm$.006} & 81\% & 5\% \\
\bottomrule
\end{tabular}
\caption{Co-optimization convergence. T1+T2 = fraction handled by cheapest tiers.}
\label{tab:coopt}
\end{table}

\subsection{Human Evaluation}
\label{sec:human_eval}

We evaluate CS Response Generation and Financial Summarization on 200 samples each, rated by 3 domain experts on factual accuracy, completeness, fluency, and helpfulness (5-point Likert), plus win/tie/loss vs.\ Always-T4.

\begin{table}[t]
	\centering
	\small
	\resizebox{\columnwidth}{!}{%
		\begin{tabular}{@{}lccccc@{}}
			\toprule
			& \multicolumn{3}{c}{\textbf{Win/Tie/Loss (\%)}} & \textbf{Likert} & \textbf{Kripp.} \\
			\cmidrule(lr){2-4}
			\textbf{Task} & \textbf{W} & \textbf{T} & \textbf{L} & \textbf{Ours / T4} & $\alpha$ \\
			\midrule
			CS Resp. & 8{\scriptsize\,$\pm$3.8} & 65{\scriptsize\,$\pm$6.6} & 27{\scriptsize\,$\pm$6.2} & 4.1 / 4.3 & 0.72 \\
			Fin.\ Summ. & 11{\scriptsize\,$\pm$4.3} & 65{\scriptsize\,$\pm$6.6} & 24{\scriptsize\,$\pm$5.9} & 4.0 / 4.2 & 0.68 \\
			\bottomrule
		\end{tabular}%
	}
	\caption{Human evaluation (200 samples, 3 annotators). $\pm$ = 95\% Wilson CI half-widths. 74.5\% of responses match or exceed T4 quality.}
	\label{tab:human_eval}
\end{table}

Averaged across tasks, 74.5\% of routed responses match or exceed frontier quality.
Among the 24--27\% rated worse, 68\% were ``slightly worse'' (Likert $\leq$1 point difference) and 32\% ``substantially worse,'' meaning ${\sim}$8--9\% of all queries received substantially degraded responses, a deployment risk requiring mitigation.
Router decisions agree with human judgment in 84--87\% of cases, with disagreements predominantly conservative (escalating unnecessarily rather than missing quality issues). We focus human evaluation on these two generation tasks because that is precisely where automated metrics are most fragile; the four structured tasks (NER, intent, clause extraction, risk assessment) are evaluated against expert annotations with high inter-annotator agreement (Financial NER F1: 0.93; Legal Clause F1: 0.91; Legal Risk Cohen's $\kappa$: 0.83; CS Intent uses public BANKING77~\citep{casanueva2020banking77}), making targeted human evaluation lower-priority for those tasks and a candidate for follow-up work rather than a gap in the present claims.

\paragraph{Failure Patterns.}
Three dominant patterns explain quality degradation. Multi-step reasoning accounts for 42\% of failures, since the router processes only the query text and not the surrounding document context. Domain-specific knowledge accounts for 31\%, typically rare instruments or recently issued regulations. The remaining 27\% are ambiguous-difficulty cases: syntactically simple queries that require nuanced generation.

\section{Deployment Experience}
\label{sec:deployment}

\textsc{RouteNLP} has been in pilot evaluation at our enterprise partner's customer service division for 8 weeks (${\sim}$5K queries/day).

\begin{table}[t]
\centering
\small
\setlength{\tabcolsep}{3pt}
\begin{tabular}{@{}lcc@{}}
\toprule
\textbf{Metric} & \textbf{Simulation} & \textbf{Pilot} \\
\midrule
Cost reduction vs.\ Always-T4 & 62\% & 58\% \\
T1 routing share & 51.1\% & 44.2\% \\
T2 routing share & 26.0\% & 27.8\% \\
T3 routing share & 15.6\% & 18.3\% \\
T4 routing share & 7.3\%\phantom{0} & 9.7\%\phantom{0} \\
Coverage violation rate & 4.2\% & 4.8\%\phantom{0} \\
\bottomrule
\end{tabular}
\caption{Simulated vs.\ pilot deployment metrics (8 weeks, ${\sim}$5K queries/day). The pilot shows higher T3/T4 usage due to more complex queries in live traffic.}
\label{tab:pilot}
\end{table}

Cost savings of 58\% are within 7\% of simulation predictions, with the gap attributable to more complex queries in live traffic (T4 usage: 9.7\% vs.\ simulated 7.3\%). Coverage violations remain within the 5\% target at 4.8\%, with weekly recalibration sufficient. For more dynamic environments where weekly recalibration may not suffice, we are implementing online threshold adaptation in the spirit of adaptive conformal methods~\citep{angelopoulos2024conformal} as part of an upcoming Phase~2 deployment. The pilot identified two novel failure patterns: OCR artifacts from scanned documents and multi-turn conversation references; both were handled conservatively via escalation.

\paragraph{Quality Audit.}
A retrospective audit by two domain experts (senior agents, 5+ years experience) on 500 random pilot responses showed 91\% acceptable, 6.4\% marginal, and 2.6\% unacceptable (vs.\ 1.8\% baseline; $\kappa\!=\!0.79$).
The partner deemed the 0.8pp increase in unacceptable responses acceptable given 58\% cost savings. Customer complaint rates showed no statistically significant change, and average first-response latency (an empirical pilot measurement, distinct from the simulated p99 reported in Table~\ref{tab:main_results}) decreased by 23\%.
The pilot was a shadow deployment with partial traffic routing (not a randomized A/B test), limiting causal attribution; a full A/B evaluation is planned for Phase~2.

\paragraph{Practical Considerations.}
The DistilBERT router adds 4.2ms per query (p99: 8.1ms), negligible relative to LLM inference. The framework is portfolio-agnostic: when frontier models change, only thresholds and quality labels need updating. Cost savings remain substantial across realistic cost ratios (58\% at 200$\times$, 41\% at 100$\times$; break-even at ${\sim}$25$\times$). During a 45-minute T4 outage in the pilot, automatic T3 rerouting incurred only 2.1\% quality degradation.
The system is most effective for heterogeneous multi-task workloads with heavy-tailed difficulty distributions; it provides limited benefit for single-task deployments, workloads dominated by hard queries, very small volumes ($<$100/day), or deployments where $>$3\% unacceptable rate is intolerable without per-query human review. Full deployment discussion is in Appendix~\ref{app:deployment_details}.

\paragraph{Lessons from the Pilot.}
Three operational lessons came out of the 8-week deployment, and they likely apply beyond customer service. First, failure modes evolved in both predictable and unpredictable ways. The three categories identified in the benchmark (multi-step reasoning, domain-specific knowledge, and ambiguous difficulty) all reappeared in pilot logs. Two new modes that the benchmark had missed also showed up: OCR artifacts from scanned documents, and multi-turn conversation references. Both were handled by escalation rather than by additional distillation, since these failure classes are not the kind that portfolio improvement addresses. Second, the choice of recalibration cadence matters. Weekly recalibration was sufficient at the drift rate we observed, with coverage violations holding at 4.8\% against the 5\% target. We do not expect weekly to suffice in more dynamic environments, which is why we are building online threshold adaptation for the Phase~2 deployment. Third, portfolio swaps turned out to be routine. Mid-pilot, we replaced T4 with GPT-4o-mini for classification queries, which compressed the effective cost ratio from $800\times$ to about $200\times$. The routing distribution adjusted on its own and no retraining was needed. This behavior is consistent with the portfolio-agnostic design and reinforces our argument that closed-loop co-optimization is a deployment-grade contribution rather than a research-only one.

\section{Conclusion}
\label{sec:conclusion}

We presented \textsc{RouteNLP}, a cost-aware routing and cascading framework validated through an 8-week pilot (58\% cost reduction, 91\% acceptance, ${\sim}$5K queries/day). The distillation-routing co-optimization loop, which integrates failure clustering with targeted distillation and automatic router retraining, provides over twice the cost reduction of random distillation. On our six-task benchmark, costs are reduced 40--85\% while retaining 96--100\% quality on structured tasks and 96--98\% on generation tasks. Human evaluation confirms 74.5\% of generation outputs match or exceed frontier quality, though 8--9\% show substantial degradation requiring mitigation.

\paragraph{Reproducibility.} All materials are available at: \url{https://github.com/bettyguo/RouteNLP}.

\section*{Acknowledgments}

We thank The University of Hong Kong and Stellaris AI Limited for their support of this work, and the anonymous reviewers for constructive feedback that improved the paper.

\section*{Limitations}

(1)~Pilot deployment covers only customer service (${\sim}$5K queries/day, 8 weeks); finance and legal claims rely on benchmark simulation. (2)~The benchmark adapts public datasets with enterprise annotations rather than proprietary data. (3)~The co-optimization loop ran on benchmark data, not production failure logs. (4)~The pilot was a shadow deployment without A/B testing. (5)~Conformal coverage degrades under distribution shift (up to 8.1\% violations vs.\ 5\% target). (6)~English-only evaluation. (7)~BERTScore proxy agreement with humans (84--87\%) is not verified under domain shift. (8)~Cost savings depend on cost structures; baseline adaptation may not fully preserve 2-model inductive biases. (9)~The co-optimization loop incurs ${\sim}$\$2,400 one-time cost at our scale.

\section*{Ethical Considerations}

Cost-aware routing creates quality disparities: queries routed to cheaper models receive less capable responses. Our framework mitigates this through task-level quality constraints and conformal calibration, but individual-level variance exists. We recommend organizations disclose model tier usage, monitor routing patterns for systematic disparities across demographic groups, and implement fairness-constrained routing that escalates segments falling below quality thresholds. The co-optimization loop (${\sim}$120 GPU-hours) yields 40--85\% ongoing inference cost reduction, a net environmental benefit. Our benchmark uses publicly available data with no PII; organizations should ensure data protection compliance when routing sensitive queries through external APIs.

\bibliography{references}

@inproceedings{aggarwal2024automix,
	author           = {Pranjal Aggarwal and
	Aman Madaan and
	Ankit Anand and
	Srividya Pranavi Potharaju and
	Swaroop Mishra and
	Pei Zhou and
	Aditya Gupta and
	Dheeraj Rajagopal and
	Karthik Kappaganthu and
	Yiming Yang and
	Shyam Upadhyay and
	Manaal Faruqui and
	Mausam},
	editor           = {Amir Globersons and
	Lester Mackey and
	Danielle Belgrave and
	Angela Fan and
	Ulrich Paquet and
	Jakub M. Tomczak and
	Cheng Zhang},
	title            = {AutoMix: Automatically Mixing Language Models},
	booktitle        = {Advances in Neural Information Processing Systems 38: Annual Conference
	on Neural Information Processing Systems 2024, NeurIPS 2024, Vancouver,
	BC, Canada, December 10 - 15, 2024},
	year             = {2024},
	bburl             = {http://papers.nips.cc/paper\_files/paper/2024/hash/ecda225cb187b40ea8edc1f46b03ffda-Abstract-Conference.html},
	bibburl           = {https://dblp.org/rec/conf/nips/AggarwalMAPMZGR24.bib},
	bibsource        = {dblp computer science bibliography, https://dblp.org},
}

@inproceedings{alvarado2015domain,
	author           = {Julio Cesar Salinas Alvarado and
	Karin Verspoor and
	Timothy Baldwin},
	editor           = {Ben Hachey and
	Kellie Webster},
	title            = {Domain Adaption of Named Entity Recognition to Support Credit Risk
	Assessment},
	booktitle        = {Proceedings of the Australasian Language Technology Association Workshop,
	{ALTA} 2015, Parramatta, Australia, December 8 - 9, 2015},
	pages            = {84--90},
	publisher        = {{ACL}},
	year             = {2015},
	bburl             = {https://aclanthology.org/U15-1010/},
	bibburl           = {https://dblp.org/rec/conf/acl-alta/AlvaradoVB15.bib},
	bibsource        = {dblp computer science bibliography, https://dblp.org},
}

@article{bates2021distribution,
	author           = {Stephen Bates and
	Anastasios Angelopoulos and
	Lihua Lei and
	Jitendra Malik and
	Michael I. Jordan},
	title            = {Distribution-free, Risk-controlling Prediction Sets},
	journal          = {J. {ACM}},
	volume           = {68},
	number           = {6},
	pages            = {43:1--43:34},
	year             = {2021},
	bburl             = {https://bdoi.org/10.1145/3478535},
	bdoi              = {10.1145/3478535},
	bibburl           = {https://dblp.org/rec/journals/jacm/BatesALMJ21.bib},
	bibsource        = {dblp computer science bibliography, https://dblp.org},
}

@inproceedings{cai2024medusa,
	author           = {Tianle Cai and
	Yuhong Li and
	Zhengyang Geng and
	Hongwu Peng and
	Jason D. Lee and
	Deming Chen and
	Tri Dao},
	editor           = {Ruslan Salakhutdinov and
	Zico Kolter and
	Katherine A. Heller and
	Adrian Weller and
	Nuria Oliver and
	Jonathan Scarlett and
	Felix Berkenkamp},
	title            = {Medusa: Simple {LLM} Inference Acceleration Framework with Multiple
	Decoding Heads},
	booktitle        = {Forty-first International Conference on Machine Learning, {ICML} 2024,
	Vienna, Austria, July 21-27, 2024},
	series           = {Proceedings of Machine Learning Research},
	volume           = {235},
	pages            = {5209--5235},
	publisher        = {{PMLR} / OpenReview.net},
	year             = {2024},
	bburl             = {https://proceedings.mlr.press/v235/cai24b.html},
	bibburl           = {https://dblp.org/rec/conf/icml/CaiLGPLCD24.bib},
	bibsource        = {dblp computer science bibliography, https://dblp.org},
}

@article{casanueva2020banking77,
	author           = {I{\~{n}}igo Casanueva and
	Tadas Temcinas and
	Daniela Gerz and
	Matthew Henderson and
	Ivan Vulic},
	title            = {Efficient Intent Detection with Dual Sentence Encoders},
	journal          = {arXiv preprint},
	volume           = {arXiv.2003.04807},
	year             = {2020},
	beprinttype       = {arXiv},
	beprint           = {2003.04807},
	bibburl           = {https://dblp.org/rec/journals/corr/abs-2003-04807.bib},
	bibsource        = {dblp computer science bibliography, https://dblp.org},
}

@inproceedings{chen2020frugalml,
	author           = {Lingjiao Chen and
	Matei Zaharia and
	James Y. Zou},
	editor           = {Hugo Larochelle and
	Marc'Aurelio Ranzato and
	Raia Hadsell and
	Maria{-}Florina Balcan and
	Hsuan{-}Tien Lin},
	title            = {FrugalML: How to use {ML} Prediction APIs more accurately and cheaply},
	booktitle        = {Advances in Neural Information Processing Systems 33: Annual Conference
	on Neural Information Processing Systems 2020, NeurIPS 2020, December
	6-12, 2020, virtual},
	year             = {2020},
	bburl             = {https://proceedings.neurips.cc/paper/2020/hash/789ba2ae4d335e8a2ad283a3f7effced-Abstract.html},
	bibburl           = {https://dblp.org/rec/conf/nips/ChenZZ20.bib},
	bibsource        = {dblp computer science bibliography, https://dblp.org},
}

@article{chen2024frugalgpt,
	author           = {Lingjiao Chen and
	Matei Zaharia and
	James Zou},
	title            = {FrugalGPT: How to Use Large Language Models While Reducing Cost and
	Improving Performance},
	journal          = {Trans. Mach. Learn. Res.},
	volume           = {2024},
	year             = {2024},
	bburl             = {https://openreview.net/forum?id=cSimKw5p6R},
	bibburl           = {https://dblp.org/rec/journals/tmlr/ChenZ024.bib},
	bibsource        = {dblp computer science bibliography, https://dblp.org},
}

@inproceedings{ding2024hybrid,
	author           = {Dujian Ding and
	Ankur Mallick and
	Chi Wang and
	Robert Sim and
	Subhabrata Mukherjee and
	Victor R{\"{u}}hle and
	Laks V. S. Lakshmanan and
	Ahmed Hassan Awadallah},
	title            = {Hybrid {LLM:} Cost-Efficient and Quality-Aware Query Routing},
	booktitle        = {The Twelfth International Conference on Learning Representations,
	{ICLR} 2024, Vienna, Austria, May 7-11, 2024},
	publisher        = {OpenReview.net},
	year             = {2024},
	bburl             = {https://openreview.net/forum?id=02f3mUtqnM},
	bibburl           = {https://dblp.org/rec/conf/iclr/DingM0SMRLA24.bib},
	bibsource        = {dblp computer science bibliography, https://dblp.org},
}

@article{frantar2023gptq,
	author           = {Elias Frantar and
	Saleh Ashkboos and
	Torsten Hoefler and
	Dan Alistarh},
	title            = {{GPTQ:} Accurate Post-Training Quantization for Generative Pre-trained
	Transformers},
	journal          = {arXiv preprint},
	volume           = {arXiv.2210.17323},
	year             = {2022},
	bburl             = {https://bdoi.org/10.48550/arXiv.2210.17323},
	bdoi              = {10.48550/ARXIV.2210.17323},
	beprinttype       = {arXiv},
	beprint           = {2210.17323},
	bibburl           = {https://dblp.org/rec/journals/corr/abs-2210-17323.bib},
	bibsource        = {dblp computer science bibliography, https://dblp.org},
}

@inproceedings{hendrycks2021cuad,
	author           = {Dan Hendrycks and
	Collin Burns and
	Anya Chen and
	Spencer Ball},
	editor           = {Joaquin Vanschoren and
	Sai{-}Kit Yeung},
	title            = {{CUAD:} An Expert-Annotated {NLP} Dataset for Legal Contract Review},
	booktitle        = {Proceedings of the Neural Information Processing Systems Track on
	Datasets and Benchmarks 1, NeurIPS Datasets and Benchmarks 2021, December
	2021, virtual},
	year             = {2021},
	bburl             = {https://datasets-benchmarks-proceedings.neurips.cc/paper/2021/hash/6ea9ab1baa0efb9e19094440c317e21b-Abstract-round1.html},
	bibburl           = {https://dblp.org/rec/conf/nips/HendrycksBCB21.bib},
	bibsource        = {dblp computer science bibliography, https://dblp.org},
}

@article{hinton2015distilling,
	author           = {Geoffrey E. Hinton and
	Oriol Vinyals and
	Jeffrey Dean},
	title            = {Distilling the Knowledge in a Neural Network},
	journal          = {arXiv preprint},
	volume           = {arXiv.1503.02531},
	year             = {2015},
	bburl             = {http://arxiv.org/abs/1503.02531},
	beprinttype       = {arXiv},
	beprint           = {1503.02531},
	bibburl           = {https://dblp.org/rec/journals/corr/HintonVD15.bib},
	bibsource        = {dblp computer science bibliography, https://dblp.org},
}

@article{jiang2024mixtral,
	author           = {Albert Q. Jiang and
	Alexandre Sablayrolles and
	Antoine Roux and
	Arthur Mensch and
	Blanche Savary and
	Chris Bamford and
	Devendra Singh Chaplot and
	Diego de Las Casas and
	Emma Bou Hanna and
	Florian Bressand and
	Gianna Lengyel and
	Guillaume Bour and
	Guillaume Lample and
	L{\'{e}}lio Renard Lavaud and
	Lucile Saulnier and
	Marie{-}Anne Lachaux and
	Pierre Stock and
	Sandeep Subramanian and
	Sophia Yang and
	Szymon Antoniak and
	Teven Le Scao and
	Th{\'{e}}ophile Gervet and
	Thibaut Lavril and
	Thomas Wang and
	Timoth{\'{e}}e Lacroix and
	William El Sayed},
	title            = {Mixtral of Experts},
	journal          = {arXiv preprint},
	volume           = {arXiv.2401.04088},
	year             = {2024},
	bburl             = {https://bdoi.org/10.48550/arXiv.2401.04088},
	bdoi              = {10.48550/ARXIV.2401.04088},
	beprinttype       = {arXiv},
	beprint           = {2401.04088},
	bibburl           = {https://dblp.org/rec/journals/corr/abs-2401-04088.bib},
	bibsource        = {dblp computer science bibliography, https://dblp.org},
}

@inproceedings{kuhn2023semantic,
	author           = {Lorenz Kuhn and
	Yarin Gal and
	Sebastian Farquhar},
	title            = {Semantic Uncertainty: Linguistic Invariances for Uncertainty Estimation
	in Natural Language Generation},
	booktitle        = {The Eleventh International Conference on Learning Representations,
	{ICLR} 2023, Kigali, Rwanda, May 1-5, 2023},
	publisher        = {OpenReview.net},
	year             = {2023},
	bburl             = {https://openreview.net/forum?id=VD-AYtP0dve},
	bibburl           = {https://dblp.org/rec/conf/iclr/KuhnGF23.bib},
	bibsource        = {dblp computer science bibliography, https://dblp.org},
}

@inproceedings{kwon2023vllm,
	author           = {Woosuk Kwon and
	Zhuohan Li and
	Siyuan Zhuang and
	Ying Sheng and
	Lianmin Zheng and
	Cody Hao Yu and
	Joseph Gonzalez and
	Hao Zhang and
	Ion Stoica},
	editor           = {Jason Flinn and
	Margo I. Seltzer and
	Peter Druschel and
	Antoine Kaufmann and
	Jonathan Mace},
	title            = {Efficient Memory Management for Large Language Model Serving with
	PagedAttention},
	booktitle        = {Proceedings of the 29th Symposium on Operating Systems Principles,
	{SOSP} 2023, Koblenz, Germany, October 23-26, 2023},
	pages            = {611--626},
	publisher        = {{ACM}},
	year             = {2023},
	bburl             = {https://bdoi.org/10.1145/3600006.3613165},
	bdoi              = {10.1145/3600006.3613165},
	bibburl           = {https://dblp.org/rec/conf/sosp/KwonLZ0ZY0ZS23.bib},
	bibsource        = {dblp computer science bibliography, https://dblp.org},
}

@inproceedings{leviathan2023speculative,
	author           = {Yaniv Leviathan and
	Matan Kalman and
	Yossi Matias},
	editor           = {Andreas Krause and
	Emma Brunskill and
	Kyunghyun Cho and
	Barbara Engelhardt and
	Sivan Sabato and
	Jonathan Scarlett},
	title            = {Fast Inference from Transformers via Speculative Decoding},
	booktitle        = {International Conference on Machine Learning, {ICML} 2023, 23-29 July
	2023, Honolulu, Hawaii, {USA}},
	series           = {Proceedings of Machine Learning Research},
	volume           = {202},
	pages            = {19274--19286},
	publisher        = {{PMLR}},
	year             = {2023},
	bburl             = {https://proceedings.mlr.press/v202/leviathan23a.html},
	bibburl           = {https://dblp.org/rec/conf/icml/LeviathanKM23.bib},
	bibsource        = {dblp computer science bibliography, https://dblp.org},
}

@inproceedings{lin2024awq,
	author           = {Ji Lin and
	Jiaming Tang and
	Haotian Tang and
	Shang Yang and
	Wei{-}Ming Chen and
	Wei{-}Chen Wang and
	Guangxuan Xiao and
	Xingyu Dang and
	Chuang Gan and
	Song Han},
	editor           = {Phillip B. Gibbons and
	Gennady Pekhimenko and
	Christopher De Sa},
	title            = {{AWQ:} Activation-aware Weight Quantization for On-Device {LLM} Compression
	and Acceleration},
	booktitle        = {Proceedings of the Seventh Annual Conference on Machine Learning and
	Systems, MLSys 2024, Santa Clara, CA, USA, May 13-16, 2024},
	publisher        = {mlsys.org},
	year             = {2024},
	bburl             = {https://proceedings.mlsys.org/paper\_files/paper/2024/hash/42a452cbafa9dd64e9ba4aa95cc1ef21-Abstract-Conference.html},
	bibburl           = {https://dblp.org/rec/conf/mlsys/0002TTYCWXDG024.bib},
	bibsource        = {dblp computer science bibliography, https://dblp.org},
}

@inproceedings{madaan2024selfrefine,
	author           = {Aman Madaan and
	Niket Tandon and
	Prakhar Gupta and
	Skyler Hallinan and
	Luyu Gao and
	Sarah Wiegreffe and
	Uri Alon and
	Nouha Dziri and
	Shrimai Prabhumoye and
	Yiming Yang and
	Shashank Gupta and
	Bodhisattwa Prasad Majumder and
	Katherine Hermann and
	Sean Welleck and
	Amir Yazdanbakhsh and
	Peter Clark},
	editor           = {Alice Oh and
	Tristan Naumann and
	Amir Globerson and
	Kate Saenko and
	Moritz Hardt and
	Sergey Levine},
	title            = {Self-Refine: Iterative Refinement with Self-Feedback},
	booktitle        = {Advances in Neural Information Processing Systems 36: Annual Conference
	on Neural Information Processing Systems 2023, NeurIPS 2023, New Orleans,
	LA, USA, December 10 - 16, 2023},
	year             = {2023},
	bburl             = {http://papers.nips.cc/paper\_files/paper/2023/hash/91edff07232fb1b55a505a9e9f6c0ff3-Abstract-Conference.html},
	bibburl           = {https://dblp.org/rec/conf/nips/MadaanTGHGW0DPY23.bib},
	bibsource        = {dblp computer science bibliography, https://dblp.org},
}

@article{ong2025routellm,
	author           = {Isaac Ong and
	Amjad Almahairi and
	Vincent Wu and
	Wei{-}Lin Chiang and
	Tianhao Wu and
	Joseph E. Gonzalez and
	M. Waleed Kadous and
	Ion Stoica},
	title            = {RouteLLM: Learning to Route LLMs with Preference Data},
	journal          = {arXiv preprint},
	volume           = {arXiv.2406.18665},
	year             = {2024},
	bburl             = {https://bdoi.org/10.48550/arXiv.2406.18665},
	bdoi              = {10.48550/ARXIV.2406.18665},
	beprinttype       = {arXiv},
	beprint           = {2406.18665},
	bibburl           = {https://dblp.org/rec/journals/corr/abs-2406-18665.bib},
	bibsource        = {dblp computer science bibliography, https://dblp.org},
}

@inproceedings{quach2024conformal,
	author           = {Victor Quach and
	Adam Fisch and
	Tal Schuster and
	Adam Yala and
	Jae Ho Sohn and
	Tommi S. Jaakkola and
	Regina Barzilay},
	title            = {Conformal Language Modeling},
	booktitle        = {The Twelfth International Conference on Learning Representations,
	{ICLR} 2024, Vienna, Austria, May 7-11, 2024},
	publisher        = {OpenReview.net},
	year             = {2024},
	bburl             = {https://openreview.net/forum?id=pzUhfQ74c5},
	bibburl           = {https://dblp.org/rec/conf/iclr/QuachFSYSJB24.bib},
	bibsource        = {dblp computer science bibliography, https://dblp.org},
}

@article{sanh2019distilbert,
	author           = {Victor Sanh and
	Lysandre Debut and
	Julien Chaumond and
	Thomas Wolf},
	title            = {DistilBERT, a distilled version of {BERT:} smaller, faster, cheaper
	and lighter},
	journal          = {arXiv preprint},
	volume           = {arXiv.1910.01108},
	year             = {2019},
	bburl             = {http://arxiv.org/abs/1910.01108},
	beprinttype       = {arXiv},
	beprint           = {1910.01108},
	bibburl           = {https://dblp.org/rec/journals/corr/abs-1910-01108.bib},
	bibsource        = {dblp computer science bibliography, https://dblp.org},
}

@inproceedings{shazeer2017moe,
	author           = {Noam Shazeer and
	Azalia Mirhoseini and
	Krzysztof Maziarz and
	Andy Davis and
	Quoc V. Le and
	Geoffrey E. Hinton and
	Jeff Dean},
	title            = {Outrageously Large Neural Networks: The Sparsely-Gated Mixture-of-Experts
	Layer},
	booktitle        = {5th International Conference on Learning Representations, {ICLR} 2017,
	Toulon, France, April 24-26, 2017, Conference Track Proceedings},
	publisher        = {OpenReview.net},
	year             = {2017},
	bburl             = {https://openreview.net/forum?id=B1ckMDqlg},
	bibburl           = {https://dblp.org/rec/conf/iclr/ShazeerMMDLHD17.bib},
	bibsource        = {dblp computer science bibliography, https://dblp.org},
}

@article{sheng2024slora,
	author           = {Ying Sheng and
	Shiyi Cao and
	Dacheng Li and
	Coleman Hooper and
	Nicholas Lee and
	Shuo Yang and
	Christopher Chou and
	Banghua Zhu and
	Lianmin Zheng and
	Kurt Keutzer and
	Joseph E. Gonzalez and
	Ion Stoica},
	title            = {S-LoRA: Serving Thousands of Concurrent LoRA Adapters},
	journal          = {arXiv preprint},
	volume           = {arXiv.2311.03285},
	year             = {2023},
	bburl             = {https://bdoi.org/10.48550/arXiv.2311.03285},
	bdoi              = {10.48550/ARXIV.2311.03285},
	beprinttype       = {arXiv},
	beprint           = {2311.03285},
	bibburl           = {https://dblp.org/rec/journals/corr/abs-2311-03285.bib},
	bibsource        = {dblp computer science bibliography, https://dblp.org},
}

@article{shnitzer2023routing,
	author           = {Tal Shnitzer and
	Anthony Ou and
	M{\'{\i}}rian Silva and
	Kate Soule and
	Yuekai Sun and
	Justin Solomon and
	Neil Thompson and
	Mikhail Yurochkin},
	title            = {Large Language Model Routing with Benchmark Datasets},
	journal          = {arXiv preprint},
	volume           = {arXiv.2309.15789},
	year             = {2023},
	bburl             = {https://bdoi.org/10.48550/arXiv.2309.15789},
	bdoi              = {10.48550/ARXIV.2309.15789},
	beprinttype       = {arXiv},
	beprint           = {2309.15789},
	bibburl           = {https://dblp.org/rec/journals/corr/abs-2309-15789.bib},
	bibsource        = {dblp computer science bibliography, https://dblp.org},
}

@inproceedings{varshney2022cascading,
	author           = {Neeraj Varshney and
	Chitta Baral},
	editor           = {Yoav Goldberg and
	Zornitsa Kozareva and
	Yue Zhang},
	title            = {Model Cascading: Towards Jointly Improving Efficiency and Accuracy
	of {NLP} Systems},
	booktitle        = {Proceedings of the 2022 Conference on Empirical Methods in Natural
	Language Processing, {EMNLP} 2022, Abu Dhabi, United Arab Emirates,
	December 7-11, 2022},
	pages            = {11007--11021},
	publisher        = {Association for Computational Linguistics},
	year             = {2022},
	bburl             = {https://bdoi.org/10.18653/v1/2022.emnlp-main.756},
	bdoi              = {10.18653/V1/2022.EMNLP-MAIN.756},
	bibburl           = {https://dblp.org/rec/conf/emnlp/VarshneyB22.bib},
	bibsource        = {dblp computer science bibliography, https://dblp.org},
}

@article{wu2023bloomberggpt,
	author           = {Shijie Wu and
	Ozan Irsoy and
	Steven Lu and
	Vadim Dabravolski and
	Mark Dredze and
	Sebastian Gehrmann and
	Prabhanjan Kambadur and
	David S. Rosenberg and
	Gideon Mann},
	title            = {BloombergGPT: {A} Large Language Model for Finance},
	journal          = {arXiv preprint},
	volume           = {arXiv.2303.17564},
	year             = {2023},
	bburl             = {https://bdoi.org/10.48550/arXiv.2303.17564},
	bdoi              = {10.48550/ARXIV.2303.17564},
	beprinttype       = {arXiv},
	beprint           = {2303.17564},
	bibburl           = {https://dblp.org/rec/journals/corr/abs-2303-17564.bib},
	bibsource        = {dblp computer science bibliography, https://dblp.org},
}

@inproceedings{yu2022orca,
	author           = {Gyeong{-}In Yu and
	Joo Seong Jeong and
	Geon{-}Woo Kim and
	Soojeong Kim and
	Byung{-}Gon Chun},
	editor           = {Marcos K. Aguilera and
	Hakim Weatherspoon},
	title            = {Orca: {A} Distributed Serving System for Transformer-Based Generative
	Models},
	booktitle        = {16th {USENIX} Symposium on Operating Systems Design and Implementation,
	{OSDI} 2022, Carlsbad, CA, USA, July 11-13, 2022},
	pages            = {521--538},
	publisher        = {{USENIX} Association},
	year             = {2022},
	bburl             = {https://www.usenix.org/conference/osdi22/presentation/yu},
	bibburl           = {https://dblp.org/rec/conf/osdi/YuJKKC22.bib},
	bibsource        = {dblp computer science bibliography, https://dblp.org},
}

@article{yue2024mixture,
	author           = {Murong Yue and
	Jie Zhao and
	Min Zhang and
	Liang Du and
	Ziyu Yao},
	title            = {Large Language Model Cascades with Mixture of Thoughts Representations
	for Cost-efficient Reasoning},
	journal          = {arXiv preprint},
	volume           = {arXiv.2310.03094},
	year             = {2023},
	beprinttype       = {arXiv},
	beprint           = {2310.03094}
}

@inproceedings{angelopoulos2024conformal,
	author           = {Vincent Blot and
	Anastasios Nikolas Angelopoulos and
	Michael I. Jordan and
	Nicolas J.{-}B. Brunel},
	editor           = {Yingzhen Li and
	Stephan Mandt and
	Shipra Agrawal and
	Mohammad Emtiyaz Khan},
	title            = {Automatically Adaptive Conformal Risk Control},
	booktitle        = {International Conference on Artificial Intelligence and Statistics,
	{AISTATS} 2025, Mai Khao, Thailand, 3-5 May 2025},
	series           = {Proceedings of Machine Learning Research},
	volume           = {258},
	pages            = {19--27},
	publisher        = {{PMLR}},
	year             = {2025},
	bburl             = {https://proceedings.mlr.press/v258/blot25a.html},
	bibburl           = {https://dblp.org/rec/conf/aistats/BlotAJB25.bib},
	bibsource        = {dblp computer science bibliography, https://dblp.org},
}

@inproceedings{gupta2024token,
	author           = {Neha Gupta and
	Harikrishna Narasimhan and
	Wittawat Jitkrittum and
	Ankit Singh Rawat and
	Aditya Krishna Menon and
	Sanjiv Kumar},
	title            = {Language Model Cascades: Token-Level Uncertainty And Beyond},
	booktitle        = {The Twelfth International Conference on Learning Representations,
	{ICLR} 2024, Vienna, Austria, May 7-11, 2024},
	publisher        = {OpenReview.net},
	year             = {2024},
	bburl             = {https://openreview.net/forum?id=KgaBScZ4VI},
	bibburl           = {https://dblp.org/rec/conf/iclr/GuptaNJRMK24.bib},
	bibsource        = {dblp computer science bibliography, https://dblp.org},
}

@article{hu2024routerbench,
	author           = {Qitian Jason Hu and
	Jacob Bieker and
	Xiuyu Li and
	Nan Jiang and
	Benjamin Keigwin and
	Gaurav Ranganath and
	Kurt Keutzer and
	Shriyash Kaustubh Upadhyay},
	title            = {RouterBench: {A} Benchmark for Multi-LLM Routing System},
	journal          = {arXiv preprint},
	volume           = {arXiv.2403.12031},
	year             = {2024},
	bburl             = {https://bdoi.org/10.48550/arXiv.2403.12031},
	bdoi              = {10.48550/ARXIV.2403.12031},
	beprinttype       = {arXiv},
	beprint           = {2403.12031},
	bibburl           = {https://dblp.org/rec/journals/corr/abs-2403-12031.bib},
	bibsource        = {dblp computer science bibliography, https://dblp.org},
}

@inproceedings{kim2016sequence,
	title            = {Sequence-Level Knowledge Distillation},
	booktitle        = {Proceedings of the 2016 Conference on Empirical Methods in Natural
	Language Processing},
	publisher        = {Association for Computational Linguistics},
	author           = {Kim, Yoon and Rush, Alexander M.},
	year             = {2016},
	pages            = {1317–1327},
}

@inproceedings{sakota2024forc,
	author           = {Marija Sakota and
	Maxime Peyrard and
	Robert West},
	editor           = {Luz Angelica Caudillo{-}Mata and
	Silvio Lattanzi and
	Andr{\'{e}}s Mu{\~{n}}oz Medina and
	Leman Akoglu and
	Aristides Gionis and
	Sergei Vassilvitskii},
	title            = {Fly-Swat or Cannon? Cost-Effective Language Model Choice via Meta-Modeling},
	booktitle        = {Proceedings of the 17th {ACM} International Conference on Web Search
	and Data Mining, {WSDM} 2024, Merida, Mexico, March 4-8, 2024},
	pages            = {606--615},
	publisher        = {{ACM}},
	year             = {2024},
	bburl             = {https://bdoi.org/10.1145/3616855.3635825},
	bdoi              = {10.1145/3616855.3635825},
	bibburl           = {https://dblp.org/rec/conf/wsdm/SakotaP024.bib},
	bibsource        = {dblp computer science bibliography, https://dblp.org},
}

@article{pan2024metallm,
	author       = {Quang H. Nguyen and
	Duy C. Hoang and
	Juliette Decugis and
	Saurav Manchanda and
	Nitesh V. Chawla and
	Khoa D. Doan},
	title        = {MetaLLM: {A} High-performant and Cost-efficient Dynamic Framework
	for Wrapping LLMs},
	journal      = {arXiv preprint},
	volume       = {arXiv.2407.10834},
	year         = {2024},
	burl          = {https://bdoi.org/10.48550/arXiv.2407.10834},
	bdoi          = {10.48550/ARXIV.2407.10834},
	beprinttype    = {arXiv},
	beprint       = {2407.10834},
	bibburl       = {https://dblp.org/rec/journals/corr/abs-2407-10834.bib},
	bibsource    = {dblp computer science bibliography, https://dblp.org}
}

@inproceedings{dekoninck2025unified,
	author       = {Jasper Dekoninck and
	Maximilian Baader and
	Martin T. Vechev},
	editor       = {Aarti Singh and
	Maryam Fazel and
	Daniel Hsu and
	Simon Lacoste{-}Julien and
	Felix Berkenkamp and
	Tegan Maharaj and
	Kiri Wagstaff and
	Jerry Zhu},
	title        = {A Unified Approach to Routing and Cascading for LLMs},
	booktitle    = {Forty-second International Conference on Machine Learning, {ICML}
	2025, Vancouver, BC, Canada, July 13-19, 2025},
	series       = {Proceedings of Machine Learning Research},
	volume       = {267},
	publisher    = {{PMLR} / OpenReview.net},
	year         = {2025},
	bburl          = {https://proceedings.mlr.press/v267/dekoninck25a.html},
	bibburl       = {https://dblp.org/rec/conf/icml/DekoninckBV25.bib},
	bibsource    = {dblp computer science bibliography, https://dblp.org}
}

@article{wang2024copilot,
	author       = {Clovis Varangot{-}Reille and
	Christophe Bouvard and
	Antoine Gourru and
	Mathieu Ciancone and
	Marion Schaeffer and
	Fran{\c{c}}ois Jacquenet},
	title        = {bdoing More with Less - Implementing Routing Strategies in Large Language
	Model-Based Systems: An Extended Survey},
	journal      = {arXiv preprint},
	volume       = {arXiv.2502.00409},
	year         = {2025},
	burl          = {https://bdoi.org/10.48550/arXiv.2502.00409},
	bdoi          = {10.48550/ARXIV.2502.00409},
	beprinttype    = {arXiv},
	beprint       = {2502.00409},
	bibburl       = {https://dblp.org/rec/journals/corr/abs-2502-00409.bib},
	bibsource    = {dblp computer science bibliography, https://dblp.org}
}

@inproceedings{zagyva2025speed,
	author       = {Libo Zhang and
	Zhaoning Zhang and
	Xubaizhou and
	Rui Li and
	Zhiliang Tian and
	Songzhu Mei and
	Dongsheng Li},
	editor       = {Christos Christodoulopoulos and
	Tanmoy Chakraborty and
	Carolyn Rose and
	Violet Peng},
	title        = {Dovetail: {A} {CPU/GPU} Heterogeneous Speculative Decoding for {LLM}
	inference},
	booktitle    = {Proceedings of the 2025 Conference on Empirical Methods in Natural
	Language Processing, {EMNLP} 2025, Suzhou, China, November 4-9, 2025},
	pages        = {17382--17395},
	publisher    = {Association for Computational Linguistics},
	year         = {2025},
	burl          = {https://bdoi.org/10.18653/v1/2025.emnlp-main.879},
	bdoi          = {10.18653/V1/2025.EMNLP-MAIN.879},
	bibburl       = {https://dblp.org/rec/conf/emnlp/ZhangZXLTML25.bib},
	bibsource    = {dblp computer science bibliography, https://dblp.org}
}

\appendix

\section{Production Deployment Scenarios}
\label{app:scenarios}

Table~\ref{tab:scenarios} summarizes representative deployment scenarios motivating \textsc{RouteNLP}'s design, derived from requirements analysis with enterprise stakeholders.

\begin{table}[ht]
\centering
\small
\setlength{\tabcolsep}{1.8pt}
\resizebox{\columnwidth}{!}{%
\begin{tabular}{@{}lllll@{}}
\toprule
\textbf{Domain} & \textbf{Org.\ Type} & \textbf{Scale} & \textbf{Constraint} & \textbf{Frontier\%} \\
\midrule
Finance & Bank/Fund & 50K q/day & Accuracy & $\sim$25\% \\
Cust.\ Svc. & Platform & 200K q/day & p99 $<$500ms & $\sim$30\% \\
Legal & Law firm & 10K q/day & Compliance & $\sim$35\% \\
\bottomrule
\end{tabular}%
}
\caption{Deployment scenarios motivating \textsc{RouteNLP}. ``Frontier\%'' = estimated fraction of queries requiring frontier model capabilities, derived from stakeholder interviews and sampling analysis.}
\label{tab:scenarios}
\end{table}

\textbf{Financial Document Processing.}
A financial services firm processes regulatory filings, earnings reports, and customer communications. Tasks include NER for entity extraction, document classification for compliance routing, and summarization. Complexity varies: extracting entities from templated SEC filings is straightforward, while summarizing novel regulatory guidance requires frontier capabilities.

\textbf{Customer Service Automation.}
A platform handles intent classification, sentiment analysis, and response generation. Over 70\% of queries are routine (order status, FAQ matching) where small models suffice; the remaining 30\% involve nuanced complaints or policy interpretation. Strict p99 latency SLAs ($<$500ms) preclude always using frontier models.

\textbf{Legal Contract Analysis.}
Legal teams perform contract review, clause extraction, and risk assessment. Standard template extraction is well-handled by fine-tuned small models; novel risk provisions in bespoke agreements demand stronger reasoning.

\section{Feature Comparison with Prior Systems}
\label{app:comparison}

\begin{table}[ht]
\centering
\small
\setlength{\tabcolsep}{1.8pt}
\begin{tabular}{@{}lcccccc@{}}
\toprule
 & \textbf{Multi-} & \textbf{Formal} & \textbf{Co-} & \textbf{SLA} & \textbf{Output} & \textbf{Tasks} \\
\textbf{System} & \textbf{Task} & \textbf{Calib.} & \textbf{Opt.} & \textbf{Aware} & \textbf{HEval} & \textbf{Eval'd} \\
\midrule
FrugalGPT & \xmark & \xmark & \xmark & \xmark & \xmark & 1 \\
Hybrid LLM & \xmark & \xmark & \xmark & \xmark & \xmark & 1 \\
RouteLLM & \xmark & \xmark & \xmark & \xmark & $\circ$\textsuperscript{a} & 1 \\
AutoMix & \xmark & \xmark & \xmark & \xmark & \xmark & 1 \\
\textsc{RouteNLP} & \cmark & \cmark & \cmark & \cmark & \cmark & 6 \\
\bottomrule
\end{tabular}
\caption{Feature comparison. $\circ$ = partial. \textsuperscript{a}RouteLLM uses human preference data for router training but does not evaluate routed outputs. Formal Calib.\ = conformal threshold calibration.}
\label{tab:comparison}
\end{table}

\section{Router Architecture and Training Details}
\label{app:router_details}

The router uses DistilBERT-base-uncased (66M parameters) with a 2-layer MLP routing head per task family. The first layer projects concatenated [CLS] + task embedding ($768 + 64 = 832$) to 256 dimensions with ReLU and 0.1 dropout. The second projects to $K=4$ logits. Training: AdamW, lr $2 \times 10^{-5}$, batch 64, 10 epochs, early stopping (patience 3). Total: ${\sim}$67M parameters, ${\sim}$45 min on A100.

\paragraph{Quality Thresholds.}
Per-task thresholds set per enterprise SLAs: F1 $\geq 0.90$ (Financial NER), ROUGE-L $\geq 0.42$ (Summarization), F1 $\geq 0.92$ (CS Intent), BERTScore $\geq 0.65$ (CS Response), F1 $\geq 0.88$ (Legal Clause), Accuracy $\geq 0.82$ (Legal Risk).

\paragraph{BERTScore as Training Proxy.}
For generation tasks, BERTScore overstates absolute quality differences (\S\ref{sec:human_eval}), but the ranking of model capabilities is preserved (queries where cheaper models score below $\tau_t$ are also those where humans prefer the frontier model), making binary labels reliable.

\section{Conformal Calibration Details}
\label{app:conformal}

Our procedure follows conformal risk control~\citep{angelopoulos2024conformal,bates2021distribution}. The nonconformity score is binary ($s_i = \mathbb{1}[q(m_k, x_i) < \tau_t]$). For each tier $k$ and task $t$:
\begin{enumerate}[nosep]
\item Compute quality labels on 500 calibration examples.
\item Partition into correctly-handled ($s_i = 0$) and failed sets.
\item Compute uncertainty scores $u(m_k, x_i)$ for all examples.
\item Set $\delta_{k,t}$ as the $\lceil (1-\alpha)(n_0+1) \rceil$-th quantile among correctly-handled examples.
\end{enumerate}

\paragraph{Calibration Set Size Sensitivity.}
Coverage violation rates (95\% Wilson CIs): 7.2\% [3.4\%, 14.4\%] at $n\!=\!100$; 5.8\% [3.4\%, 9.6\%] at $n\!=\!250$; 4.2\% [2.5\%, 6.6\%] at $n\!=\!500$; 3.9\% [2.7\%, 5.5\%] at $n\!=\!1000$. At $n\!=\!500$, the CI upper bound marginally exceeds 5\%; we use 500 as a practical compromise.

\section{Baseline Adaptation Protocol}
\label{app:baselines}

Hybrid LLM and RouteLLM were designed for 2-model routing. We evaluate two adaptations:
\begin{itemize}[nosep]
\item \textbf{Binary-T2/T4}: Faithful to original design, routing between T2 and T4 only.
\item \textbf{Extended-4-Tier}: Replace binary head with 4-class head, train on \textsc{RouteNLP}'s data with original loss.
\end{itemize}
We report the better result (Extended-4-Tier for both; Binary incurred 2.1--3.4$\times$ higher costs). FrugalGPT's cascade naturally extends to 4 tiers. AutoMix's POMDP is reformulated with 4 actions. All baselines received identical training data, portfolio access, and calibration sets.

A hierarchical binary adaptation (T1-vs-rest $\to$ T2-vs-rest $\to$ T3-vs-T4) was not evaluated as it introduces sequential latency overhead conflating adaptation effects with architectural changes.

\section{Benchmark Data Provenance}
\label{app:benchmark_details}

\begin{itemize}[nosep]
\item \textbf{Financial NER}: Entity annotations on SEC EDGAR 10-K/10-Q filings by two NLP researchers with 3+ years financial text experience (inter-annotator F1: 0.93), following \citet{alvarado2015domain}. Annotators have NLP expertise but are not licensed financial analysts.
\item \textbf{Financial Summarization}: Earnings call transcripts from public SEC filings with expert-written reference summaries from financial analysts.
\item \textbf{CS Intent Classification}: BANKING77~\citep{casanueva2020banking77} augmented with 3,000 enterprise-specific intents from anonymized partner logs.
\item \textbf{CS Response Generation}: Queries from augmented intent dataset with reference responses by 3 experienced agents (Krippendorff's $\alpha = 0.78$).
\item \textbf{Legal Clause Extraction}: CUAD~\citep{hendrycks2021cuad} filtered to 10 most common clause types, re-annotated with span-level labels by two legal annotators (inter-annotator F1: 0.91).
\item \textbf{Legal Risk Assessment}: Novel annotations on CUAD contracts by two legal professionals (Cohen's $\kappa = 0.83$).
\end{itemize}

The benchmark adapts public academic datasets with enterprise-specific annotations, enabling reproducibility while capturing domain difficulty patterns.

\section{Co-Optimization Loop Details}
\label{app:coopt_details}

\begin{figure}[ht]
\centering
\begin{tikzpicture}[
    node distance=1.0cm,
    box/.style={rectangle, draw, rounded corners=2pt, minimum height=0.6cm, font=\scriptsize, align=center, minimum width=1.6cm},
    arrow/.style={-{Stealth[length=1.5mm]}, thick, color=purple!70},
]
\node[box, fill=orange!15] (deploy) {Deploy\\Router};
\node[box, fill=red!10, right=0.7cm of deploy] (collect) {Collect\\Escalation Logs};
\node[box, fill=yellow!15, below=0.6cm of collect] (cluster) {Cluster\\Failures};
\node[box, fill=green!15, left=0.7cm of cluster] (distill) {Targeted\\Distillation};
\node[box, fill=blue!10, left=0.7cm of distill] (update) {Update\\Models};
\node[box, fill=purple!10, above=0.6cm of update] (retrain) {Retrain\\Router};
\draw[arrow] (deploy) -- (collect);
\draw[arrow] (collect) -- (cluster);
\draw[arrow] (cluster) -- (distill);
\draw[arrow] (distill) -- (update);
\draw[arrow] (update) -- (retrain);
\draw[arrow] (retrain) -- (deploy);
\end{tikzpicture}
\caption{The co-optimization feedback loop.}
\label{fig:coopt_loop}
\end{figure}
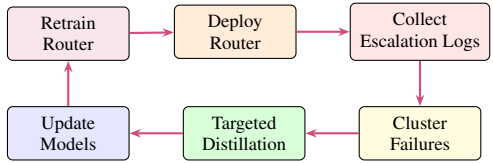

\paragraph{Failure Clustering.}
Figure~\ref{fig:coopt_loop} summarizes the loop. Router penultimate-layer representations ($\mathbb{R}^{768}$) are projected via PCA to 128-d (retaining $>$92\% variance), then $k$-means with $k\!=\!10$ (selected by silhouette score: 0.31 vs.\ 0.28 at $k\!=\!5$, 0.24 at $k\!=\!20$). Clusters ranked by size $\times$ average quality gap.

\paragraph{Distillation Data Selection.}
From top-5 clusters per task: top 30\% hardest failures (by quality gap) + 20\% random in-distribution. Frontier model generates reference outputs as teacher signals for SeqKD~\citep{kim2016sequence}.

\paragraph{Convergence.}
Across 5 seeds, the loop converged in 3 iterations (4 seeds) or 4 iterations (1 seed). Tested $\varepsilon \in \{0.001, 0.005, 0.01\}$: at 0.01, 2 iterations with 3\% higher cost; at 0.001, 5 iterations with $<$0.002 additional reduction.

\begin{table}[ht]
\centering
\small
\begin{tabular}{@{}lcccc@{}}
\toprule
\textbf{Iter.} & \textbf{Quality$\uparrow$} & \textbf{Cost$\downarrow$} & \textbf{T1+T2} & \textbf{T4} \\
\midrule
0 (init.) & .961\tiny{$\pm$.005} & .203\tiny{$\pm$.008} & 68\% & 11\% \\
1 & .964\tiny{$\pm$.004} & .178\tiny{$\pm$.007} & 74\% & 8\% \\
2 & .969\tiny{$\pm$.004} & .163\tiny{$\pm$.006} & 79\% & 6\% \\
3 (final) & .971\tiny{$\pm$.004} & .159\tiny{$\pm$.006} & 81\% & 5\% \\
\bottomrule
\end{tabular}
\caption{Co-optimization convergence. Each iteration shifts more queries to cheaper tiers.}
\label{tab:coopt_app}
\end{table}

\paragraph{Random vs.\ Targeted Distillation.}
Random distillation (same data volume, no failure clustering) reduces cost ratio from 0.203 to 0.184 (9.4\%); targeted reduces to 0.159 (21.7\%), over twice the improvement, confirming failure clustering's value.

\section{Hyperparameter Sensitivity}
\label{app:sensitivity}

\paragraph{Loss Weights.}
Sweeping $\lambda_c \in \{0.1, 0.2, 0.3, 0.5\}$ and $\lambda_q \in \{0.3, 0.5, 0.7\}$: best tradeoff at $\lambda_c = 0.3, \lambda_q = 0.5$. Higher $\lambda_q$ (0.7) improves quality by 0.004 but increases cost by 12\%. Higher $\lambda_c$ (0.5) reduces cost by 8\% but decreases quality by 0.008.

\begin{figure}[ht]
	\centering
	\begin{tikzpicture}
		\begin{axis}[
			width=0.42\textwidth,
			height=3.0cm,
			xlabel={Conformal error rate $\alpha$},
			ylabel near ticks,
			axis y line*=left,
			ylabel={Quality Ratio},
			ymin=0.94, ymax=1.0,
			xmin=0.01, xmax=0.15,
			xtick={0.01,0.03,0.05,0.10,0.15},
			xticklabel style={/pgf/number format/fixed},
			legend style={at={(0.98,0.98)}, anchor=north east, font=\tiny,
				fill=none, draw=none,
				legend columns=2, column sep=4pt},
			tick label style={font=\scriptsize},
			label style={font=\scriptsize},
			mark size=1.5pt,
			every axis plot/.append style={line width=0.8pt},
			]
			\addplot[color=blue, mark=*, thick] coordinates {
				(0.01, 0.983) (0.03, 0.977) (0.05, 0.971) (0.10, 0.961) (0.15, 0.948)
			};
			\addlegendentry{Quality}
			\addplot[color=red, mark=triangle*, thick, dashed] coordinates {
				(0.01, 0.298) (0.03, 0.218) (0.05, 0.159) (0.10, 0.112) (0.15, 0.089)
			};
			\addlegendentry{Cost}
		\end{axis}
		\begin{axis}[
			width=0.42\textwidth,
			height=3.0cm,
			axis y line*=right,
			axis x line=none,
			ylabel={Cost Ratio},
			ylabel near ticks,
			ymin=0.05, ymax=0.35,
			xmin=0.01, xmax=0.15,
			xticklabel style={/pgf/number format/fixed},
			tick label style={font=\scriptsize},
			label style={font=\scriptsize},
			mark size=1.5pt,
			]
		\end{axis}
	\end{tikzpicture}
	\caption{Effect of conformal error rate $\alpha$ on quality and cost. $\alpha\!=\!0.05$ balances both.}
	\label{fig:sensitivity}
\end{figure}
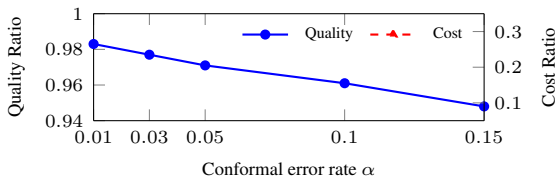

\paragraph{Conformal Error Rate.}
Figure~\ref{fig:sensitivity} shows the cost--quality tradeoff across $\alpha$ values.

\section{Extended Results}
\label{app:extended_results}

\paragraph{Per-Task Quality.}

\begin{table}[ht]
\centering
\small
\setlength{\tabcolsep}{3pt}
\begin{tabular}{@{}lccccc@{}}
\toprule
\textbf{Task} & \textbf{Metric} & \textbf{T4} & \textbf{Ours} & \textbf{Retain} & \textbf{Cost$\downarrow$}\\
\midrule
Fin.\ NER & F1 & 94.2 & 93.8\tiny{$\pm$.3} & 99.6\% & 82\% \\
Fin.\ Summ. & R-L & 48.7 & 46.9\tiny{$\pm$.4} & 96.3\% & 47\% \\
CS Intent & F1 & 96.1 & 95.8\tiny{$\pm$.2} & 99.7\% & 85\% \\
CS Resp. & BS & 72.4 & 69.7\tiny{$\pm$.6} & 96.3\% & 42\% \\
Legal Cl. & F1 & 91.6 & 90.9\tiny{$\pm$.3} & 99.2\% & 78\% \\
Legal Risk & Acc & 88.3 & 86.1\tiny{$\pm$.5} & 97.5\% & 40\% \\
\bottomrule
\end{tabular}
\caption{Per-task quality. R-L = ROUGE-L, BS = BERTScore.}
\label{tab:per_task_app}
\end{table}

\paragraph{Routing Distribution.}

\begin{table}[ht]
\centering
\small
\begin{tabular}{@{}lrrrr@{}}
\toprule
\textbf{Task} & \textbf{T1} & \textbf{T2} & \textbf{T3} & \textbf{T4} \\
\midrule
Fin.\ NER & 68\% & 22\% & 7\% & 3\% \\
Fin.\ Summ. & 31\% & 34\% & 23\% & 12\% \\
CS Intent & 72\% & 18\% & 6\% & 4\% \\
CS Response & 28\% & 30\% & 27\% & 15\% \\
Legal Clause & 62\% & 24\% & 9\% & 5\% \\
Legal Risk & 35\% & 28\% & 22\% & 15\% \\
\bottomrule
\end{tabular}
\caption{Query routing distribution across tiers.}
\label{tab:routing_dist}
\end{table}

\paragraph{Quality Threshold Sensitivity.}

\begin{table}[ht]
\centering
\small
\begin{tabular}{@{}lcccc@{}}
\toprule
\textbf{$\tau_t$ adj.} & \textbf{Quality$\uparrow$} & \textbf{Cost$\downarrow$} & \textbf{Savings} & \textbf{T4 \%} \\
\midrule
$-$10\% & .959\tiny{$\pm$.005} & .113\tiny{$\pm$.005} & 89\% & 2\% \\
$-$5\% & .965\tiny{$\pm$.004} & .132\tiny{$\pm$.006} & 87\% & 3\% \\
Baseline & .971\tiny{$\pm$.004} & .159\tiny{$\pm$.006} & 84\% & 5\% \\
$+$5\% & .978\tiny{$\pm$.003} & .218\tiny{$\pm$.008} & 78\% & 9\% \\
$+$10\% & .984\tiny{$\pm$.003} & .312\tiny{$\pm$.010} & 69\% & 14\% \\
\bottomrule
\end{tabular}
\caption{Threshold sensitivity. Savings remain $\geq$69\% even at +10\%.}
\label{tab:threshold_sens}
\end{table}

\paragraph{Robustness to Distribution Shift.}

\begin{table}[ht]
\centering
\small
\begin{tabular}{@{}lccc@{}}
\toprule
\textbf{Shift} & \textbf{Quality} & \textbf{Cost} & \textbf{Cov.\ Viol.} \\
\midrule
No shift & .971 & .159 & 4.2\% \\
Difficulty shift & .963 & .214 & 6.8\% \\
Domain shift (20\%) & .958 & .248 & 8.1\% \\
Task mix shift & .967 & .192 & 5.4\% \\
\bottomrule
\end{tabular}
\caption{Robustness under distribution shift. Target coverage violation: $\leq$5\%.}
\label{tab:robustness}
\end{table}

Under domain shift, coverage violations exceed the 5\% target, indicating recalibration is needed. The BERTScore proxy used for generation routing may also become less reliable under shift; while validated on in-distribution data (84--87\% agreement), this has not been verified under shifted distributions.

\paragraph{Baseline Advantage Across Thresholds.}
\textsc{RouteNLP} maintains the lowest cost ratio across all threshold settings: at +5\% stricter thresholds, \textsc{RouteNLP}'s cost (0.218) remains below RouteLLM (0.308) and Hybrid LLM (0.381).

\paragraph{Per-Task Detailed Comparison.}

\begin{table}[t]
	\centering
	\small
	\resizebox{\columnwidth}{!}{%
		\begin{tabular}{@{}lcccccc@{}}
			\toprule
			& \textbf{Fin.~NER} & \textbf{Fin.~Summ.} & \textbf{CS Int.} & \textbf{CS Resp.} & \textbf{Legal Cl.} & \textbf{Legal Risk} \\
			\textbf{System} & F1 & R-L & F1 & BS & F1 & Acc \\
			\midrule
			Always-T4 & 94.2 & 48.7 & 96.1 & 72.4 & 91.6 & 88.3 \\
			Always-T2 & 89.1 & 42.3 & 93.4 & 61.8 & 84.7 & 76.2 \\
			FrugalGPT & 92.8{\scriptsize\,$\pm$.3} & 47.1{\scriptsize\,$\pm$.4} & 95.3{\scriptsize\,$\pm$.2} & 69.5{\scriptsize\,$\pm$.5} & 89.8{\scriptsize\,$\pm$.3} & 85.1{\scriptsize\,$\pm$.4} \\
			Hybrid LLM & 93.1{\scriptsize\,$\pm$.4} & \textbf{47.4}{\scriptsize\,$\pm$.5} & 95.6{\scriptsize\,$\pm$.2} & \textbf{70.2}{\scriptsize\,$\pm$.6} & 90.1{\scriptsize\,$\pm$.4} & 85.7{\scriptsize\,$\pm$.5} \\
			RouteLLM & 92.9{\scriptsize\,$\pm$.3} & 47.2{\scriptsize\,$\pm$.4} & 95.4{\scriptsize\,$\pm$.2} & 69.8{\scriptsize\,$\pm$.5} & 89.9{\scriptsize\,$\pm$.3} & 85.4{\scriptsize\,$\pm$.4} \\
			AutoMix & 91.7{\scriptsize\,$\pm$.5} & 46.5{\scriptsize\,$\pm$.5} & 94.8{\scriptsize\,$\pm$.3} & 68.4{\scriptsize\,$\pm$.6} & 88.6{\scriptsize\,$\pm$.4} & 83.9{\scriptsize\,$\pm$.6} \\
			\textsc{RouteNLP} & \textbf{93.8}{\scriptsize\,$\pm$.3} & 46.9{\scriptsize\,$\pm$.4} & \textbf{95.8}{\scriptsize\,$\pm$.2} & 69.7{\scriptsize\,$\pm$.6} & \textbf{90.9}{\scriptsize\,$\pm$.3} & \textbf{86.1}{\scriptsize\,$\pm$.5} \\
			\bottomrule
		\end{tabular}%
	}
	\caption{Per-task quality across all systems (std over 5 seeds).}
	\label{tab:detailed_all}
\end{table}
\section{Latency Analysis Under Load}
\label{app:latency}

We simulate production traffic with Poisson arrivals (1K--20K queries/min) using an M/M/c queuing model. Server capacity: T1 has 8 workers (ONNX Runtime, CPU), T2/T3 have 4 workers each (vLLM on A100 GPUs), T4 is rate-limited at 60 req/s. At 10K queries/min, \textsc{RouteNLP} maintains p99 of 387ms (vs.\ 1,847ms for Always-T4) because 81\% of queries are handled by T1/T2 with sub-100ms inference. The cascade adds 15--45ms for escalated queries. The M/M/c model assumes exponential service times, which may not hold for LLM generation; SLA violation estimates may be optimistic.

\section{Extended Deployment Details}
\label{app:deployment_details}

\paragraph{Pilot Per-Task Breakdown.}
CS traffic: intent classification (${\sim}$62\%), response generation (${\sim}$28\%), sentiment analysis (${\sim}$10\%). Intent classification routed 71\% to T1 (comparable to 72\% benchmark); response generation routed only 22\% to T1 (vs.\ 28\% benchmark), suggesting generation difficulty is underestimated by the benchmark.

\paragraph{Quality Audit Details.}

\begin{table}[ht]
\centering
\small
\setlength{\tabcolsep}{3pt}
\begin{tabular}{@{}lcc@{}}
\toprule
\textbf{Quality Metric} & \textbf{RouteNLP} & \textbf{Previous} \\
\midrule
Acceptable rate & 91.0\% & 93.8\% \\
Marginal rate & 6.4\% & 4.4\% \\
Unacceptable rate & 2.6\% & 1.8\% \\
Escalation-to-human rate & 4.2\% & 3.1\% \\
Customer complaint rate & No sig.\ change & Baseline \\
\bottomrule
\end{tabular}
\caption{Pilot quality audit (500 samples, 2 raters, $\kappa\!=\!0.79$).}
\label{tab:pilot_quality}
\end{table}

\paragraph{Portfolio Evolution.}
The framework is portfolio-agnostic. During the pilot, replacing T4 with GPT-4o-mini for classification compressed the cost ratio from 800$\times$ to ${\sim}$200$\times$; routing distributions adjusted automatically (T1+T2 share: 81\%$\to$74\%). Cost savings at different ratios: 84\% at 800$\times$, 72\% at 400$\times$, 58\% at 200$\times$, 41\% at 100$\times$, 29\% at 50$\times$. Break-even at ${\sim}$25$\times$.

\paragraph{System Resilience.}
During a 45-minute T4 outage, automatic T3 rerouting incurred only 2.1\% quality degradation. Fallback: T3 serves as ceiling with relaxed thresholds; if T2/T3 down, T1 handles structured tasks and T4 handles generation.

\paragraph{Monitoring Recommendations.}
Monitor: per-tier routing rates (shifts indicate distribution change), per-task escalation rates (increases suggest model degradation), quality drift via periodic sampling, and SLA violations. Alert when escalation increases $>$10\% relative to baseline.

\paragraph{Co-Optimization on Production Data.}
The loop ran on benchmark data; pilot failure modes (OCR artifacts, multi-turn references) were not in benchmark clusters, suggesting partial mismatch. However, the three dominant benchmark failure categories were also observed in pilot logs. Running on production data is planned (estimated \$1,200/iteration).

\paragraph{Extension to Agentic Workloads.}
Reasoning models and agentic workflows introduce chain-level routing challenges. The co-optimization loop could extend to chain-level failure patterns, but this requires different quality estimation and is left to future work.

\paragraph{Infrastructure Integration.}
T1 via ONNX Runtime, T2/T3 via vLLM with LoRA adapters~\citep{sheng2024slora}, T4 via API gateway. Continuous batching~\citep{yu2022orca} enabled for self-hosted tiers. Router served as lightweight sidecar.

\paragraph{Cost Model.}
Fixed per-query pricing is used. Self-hosted costs depend on GPU utilization and infrastructure amortization; API costs change over time. Routing decisions are driven by quality thresholds and uncertainty, not absolute costs; pricing changes affect savings magnitude but not quality retention.

\end{document}